\documentclass{article}

\usepackage[final,dblblindworkshop,nonatbib]{neurips_2026}
\usepackage[backend=biber]{biblatex}

\usepackage{graphicx}
\usepackage[utf8]{inputenc} % allow utf-8 input
\usepackage[T1]{fontenc}    % use 8-bit T1 fonts
\usepackage{hyperref}       % hyperlinks
\usepackage{url}            % simple URL typesetting
\usepackage{booktabs}       % professional-quality tables
\usepackage{amsfonts}       % blackboard math symbols
\usepackage{nicefrac}       % compact symbols for 1/2, etc.
\usepackage{microtype}      % microtypography
\usepackage{xcolor}         % colors
\usepackage{comment}
\usepackage{makecell}
\usepackage{amsmath}
\usepackage[capitalize]{cleveref}
\hypersetup{hidelinks}

\usepackage{tcolorbox}
\usepackage{listings}
\tcbuselibrary{listings, skins, breakable}

\newcommand{\meanerr}[2]{%
    \makecell{$#1$ \\[-2pt] $\pm\,#2$}%
}
\newcommand{\bestmeanerr}[2]{%
    \makecell{$\mathbf{#1}$ \\[-2pt]
    $\boldsymbol{\pm}\,\mathbf{#2}$}%
}

\newtcblisting{promptbox}[1][]{
  colback=gray!5,
  colframe=gray!40,
  arc=2mm,
  boxrule=0.8pt,
  fontupper=\small\ttfamily,
  breakable,
  listing only,
  listing options={
    basicstyle=\small\ttfamily,
    breaklines=true,
    breakatwhitespace=true,
    columns=fullflexible,
    keepspaces=true
  },
  title={#1}
}

\title{AutoNovo: Agentic Program Search Discovers Transferable Nucleic Acid Design Algorithms}
\workshoptitle{Agentic AI for Biological Discovery}

\author{
  Joel Shor \\
  AI Bio Design, Allen Institute \& Move37 Labs\\
  \href{mailto:joel.shor@alleninstitute.org}{\texttt{joel.shor@alleninstitute.org}}
  \And
  Subhashini Venugopalan \\
  Google Research \\
  \href{mailto:vsubhashini@google.com}{\texttt{vsubhashini@google.com}}
}

\begin{document}

\maketitle

\begin{abstract}
  In-silico optimization of nucleic acids relies on predictive oracles that imperfectly capture biological function, so sequences optimized against one model may exploit model-specific artifacts and fail to generalize to independent models. Recent studies have shown that not all designers generalize equally well to independent models. In this work, we use agentic program search to iteratively propose, execute, evaluate, and revise code for nucleic-acid designers. We describe the successful strategies used to prompt the search. AutoNovo designers achieved the best observed mean among the evaluated methods on 3/3 DNA-design tasks in terms of in-silico performance, and 6/6 transfer tasks in terms of independent models. The designers come from a family of discovered nucleic acid designers rather than a single universal best designer. We plan to release the resulting designers. These results are the first steps in creating nucleic acid designers that maximize cross-model generalization, and thus improving the odds that in-silico designed sequences reflect real biology.
\end{abstract}

\section{Introduction}

Drug developers have been attempting to improve the therapy development process by computationally designing nucleic acids (in-silico design), which is a significantly faster and cheaper method than running large screens in a wetlab. One way of performing in-silico design of nucleic acids is to use a predictive sequence-to-function model (the oracle), that takes a sequence as input and predicts a scalar property such as translational efficiency, stability, or cell-type specific expression, and optimize the input sequence to maximize or minimize the property using a designer algorithm. A major hurdle to this in-silico design process is that the resulting synthetic sequences are often predicted to have excellent properties, when they actually do not perform well in real biological systems~\parencite{Renz2019FailureModes}.

It is expensive and time consuming to synthesize synthetic sequences in a wetlab to validate their performance in real biological systems. A cheaper and faster technique used in the field is to validate synthetic sequences using ``orthogonal validation.'' These models are trained to predict the same property, but differ in their training data, model architecture, or both. Note that this orthogonal validation is akin to a one-way test: agreement between the oracle and an orthogonal model does not guarantee that a synthetic sequence will perform well in real biological systems, but disagreement often indicates that the synthetic sequence is more an artifact of the oracle model rather than representing some aspect of biology.

A recent benchmark of nucleic acid design algorithms found empirically that designers differ significantly in their ability to agree with orthogonal models~\parencite{Shor2025NucleoBench}. In this paper, our central question is: \textbf{can agentic program search discover nucleic acid design algorithms that outperform human-designed baselines}, with an eye towards generating synthetic sequences that perform well on oracles not used during program search? Such designers, we hope, would create sequences that were significantly more likely to perform as predicted in real biological systems, and therefore be more useful for real-world applications such as therapy design.

AI code tools are a powerful technique for designing algorithms. Recent works such as AlphaEvolve~\cite{novikov2025alphaevolve}, Empirical Research Assistant~\cite{Brenner2026ERA}, SkyDiscover~\cite{liu2026skydiscover} use evolutionary or tree-based methods to iteratively improve code generated by LLMs to optimize a given objective function and demonstrate it on a range of mathematical, scientific, and engineering problems. The ERA system closes an automated loop around code generation, execution, oracle evaluation, and program revision.

In this work, we use ERA to create designers and term our approach AutoNovo. Specifically, each ERA run searches for designer programs using an objective comprising one or two optimization oracles and, in most runs, an orthogonal evaluation oracle hidden from the designer, evaluated on a small set of start sequences. The use of LLM-based evolutionary algorithms like ERA introduces the concern that an AI-generated designer might overfit to the tasks it was evaluated against and not generalize to others. We take a two-pronged approach to address this. First, we prompt it to (re-)implement specific hand-crafted designers providing high-level information of the algorithmic approach. The prompt also includes some notes (see Appendix) on generalizing to other tasks (\textbf{task generalization}).  Second, the optimization function is set to design on a specific task, and evaluate both on the task it designs on and another orthogonal task.  All tasks are chosen from the recent NucleoBench~\cite{Shor2025NucleoBench} benchmark.

Our experiments indicate that AutoNovo-generated designers against NucleoBench tasks show successful optimization and favorable task generalization. The AutoNovo designers achieved the best observed mean performances even compared with human-designed baselines on 3/3 DNA design tasks, and 6/6 cross-oracle transfer tasks (design using model X, evaluate designed sequence on model Y). The most performant AutoNovo candidates are not the same across all tasks, motivating an analysis of what distinct algorithms the agent discovered, and future work that will discover a single designer superior in all contexts.

\section{AutoNovo: Agentic program search for sequence designers}

Given a problem description, a seed implementation such as the function signature, and a scorable objective, code optimization systems like AlphaEvolve~\cite{novikov2025alphaevolve} and ERA~\cite{Brenner2026ERA} rewrite candidate programs and evaluate them to iteratively make improvements. ERA uses a tree search with an upper-confidence-bound rule which decides candidates to expand and modify. Each rewrite is executed end-to-end; the resulting metric is read from the run, and search continues toward programs that improve it. The output is ordinary, inspectable code. In summary, the agent searches over programs that implement designers, where each resulting designer searches over biological sequences. AutoNovo therefore contains two nested optimization loops: ERA searches over programs that implement designers, while each generated designer searches over biological sequences.

In AutoNovo, we prompt ERA to implement a \texttt{Designer} with a fixed interface: initialization stores a queryable oracle \texttt{model\_fn} and a length-preserving ACGT start sequence; \texttt{run(allowed\_runtime\_secs)} iteratively edits the sequence under a wall-clock budget; and \texttt{get\_samples} returns the best sequence found.  
We specifically task ERA with reimplementing \textbf{existing} hand-crafted algorithms to optimize for a task in NucleoBench~\cite{Shor2025NucleoBench}.  
We seed ERA with descriptions of four existing hand-crafted algorithms: GrAdaBeam~\parencite{Shor2025NucleoBench}, FastSeqProp~\parencite{fastseqprop}, Ledidi~\parencite{ledidi}, and Proximal Exploration~\parencite{pex}. We had an LLM implement an initial version of each algorithm twice, giving us 2 seeds for each. We then tasked ERA with improving the implementations based on the description. ERA is able to make changes, execute and evaluate the code it writes. 

In our initial experiments, we had ERA train and evaluate on a single task. For the 4 algorithms, with 2 seeds, we had 8 implementations. ERA based reimplementations of FastSeqProp and Ledidi were more performant than NucleoBench baseline implementations. The reimplementations of GrAdaBeam was not performant. On stack ranking them, we picked both seeds of FastSeqProp, one seed of Ledidi, and one seed of PEX where ERA was able to show substantial improvements.

Using the selected seeds, we ran five ERA searches. In four of them, the generated designer optimized sequences against one oracle, while ERA evaluated candidate programs using that oracle and one additional orthogonal oracle; the third oracle in the panel was used by neither and was reserved as a fully held-out evaluation task. The fifth uses both BPNet ATAC and ChromBPNet for optimization and scores both, leaving Enformer held out from the designer and from ERA. \Cref{tab:search_oracle_pairs} shows the five configurations. The final evaluation therefore contains sixteen AutoNovo designers: four from each of two FastSeqProp code seeds and four from one Ledidi code seed under the single-oracle searches, plus four from the search against both BPNet ATAC and ChromBPNet.

In each ERA search, the generated designer receives one or both of two optimization oracles as $f_{\mathrm{opt}}$. The first is the NucleoBench BPNet-lite ATAC model~\parencite{bpnet-lite,Shor2025NucleoBench}, a dilated residual CNN~\parencite{bpnet} trained on ENCODE K562 ATAC-seq~\parencite{encode}. The second is ChromBPNet~\parencite{Pampari2024ChromBPNet}, a fully convolutional network of dilated residual layers trained on K562 ATAC-seq (ENCSR483RKN). The generated designer queries $f_{\mathrm{opt}}$ through \texttt{model\_fn}. When multiple optimization oracles are used, $f_{\mathrm{opt}}$ is the sum of their scores. ERA scores the sequences produced by each candidate using $f_{\mathrm{eval}} = f_{\mathrm{opt}} + \lambda f_{\mathrm{orth}}$, where $f_{\mathrm{orth}}$ is an orthogonal evaluation score hidden from the designer when present, and is set to zero otherwise. We use the five ordered oracle configurations shown in \cref{tab:search_oracle_pairs}. The prompt also instructs the designer to generalize to other tasks and includes observations from prior work~\cite{Shor2025NucleoBench} (see the base prompt in the Appendix).

\begin{table}[htbp]
\centering
\caption{Oracle configurations used during ERA program search. The designer can query $f_{\mathrm{opt}}$, the sum of the listed optimization-oracle scores. The orthogonal oracle $f_{\mathrm{orth}}$ is hidden from the designer when present.}
\label{tab:search_oracle_pairs}
\renewcommand{\arraystretch}{1.2}
\begin{tabular}{llc}
\toprule
\textbf{Optimization oracle $f_{\mathrm{opt}}$}
& \textbf{Orthogonal oracle $f_{\mathrm{orth}}$}
& $\boldsymbol{\lambda}$ \\
\midrule
BPNet ATAC  & ChromBPNet & $1$ \\
ChromBPNet  & BPNet ATAC & $1$ \\
BPNet ATAC  & Enformer   & $0.01$ \\
ChromBPNet  & Enformer   & $0.01$ \\
BPNet ATAC + ChromBPNet & none & $1$ \\
\bottomrule
\end{tabular}
\end{table}

During program search, each candidate designer receives $f_{\mathrm{opt}}$ as \texttt{model\_fn} and is given $T=30$ minutes to design a sequence from each of three NucleoBench start sequences. ERA minimizes the mean candidate metric
\[
    f_{\mathrm{eval}} = \frac{1}{N}\sum_{i=1}^{N}
    \left[
        f_{\mathrm{opt}}(x_i)
        + \lambda f_{\mathrm{orth}}(x_i)
    \right],
    \qquad N=3,
\]
where $x_i$ is the sequence returned for start sequence $i$. We use $\lambda=1$ when $f_{\mathrm{orth}}$ is BPNet ATAC or ChromBPNet and $\lambda=0.01$ when it is Enformer, to account for differences in score scale. Crashes, use of \texttt{try}/\texttt{except}, and runs exceeding the time budget by more than a five-second grace period receive a large penalty. ERA uses this metric to decide which programs to retain and expand in its search tree.

After program search, we freeze each selected \texttt{Designer} program and make no further code changes. We then evaluate two forms of generalization. First, running a frozen program with an oracle that was not its optimization oracle $f_{\mathrm{opt}}$ during ERA search measures whether the discovered algorithm transfers to a new design task. Second, scoring a designed sequence with an oracle other than the one used to design it measures cross-oracle sequence transfer: whether the improvement persists under a different predictive model rather than reflecting exploitation of the design oracle.

\section{Tasks}\label{sec:tasks}

We evaluate designers across three DNA oracles that predict chromatin accessibility in K562: BPNet ATAC, ChromBPNet, and Enformer. During program search, ERA evaluates candidates using one of the five ordered pairs of optimization and orthogonal oracles in \cref{tab:search_oracle_pairs}. In four configurations the generated designer can query only the optimization oracle $f_{\mathrm{opt}}$, while ERA computes the candidate metric $f_{\mathrm{eval}}$ from $f_{\mathrm{opt}}$ and the orthogonal oracle $f_{\mathrm{orth}}$, which is hidden from the designer. At test time, each selected program is frozen and instantiated with each of the three oracles, which we call the \emph{design oracle}. Each resulting sequence is then scored by all three \emph{evaluation oracles}. A score is \emph{direct} when the design and evaluation oracles are the same and a \emph{transfer score} when they differ. Applying a frozen program to an oracle that was not its search-time $f_{\mathrm{opt}}$ or $f_{\mathrm{orth}}$ measures algorithm-level generalization, while scoring its output with a different oracle measures cross-oracle sequence transfer. Each AutoNovo designer has one completely held-out task that measures algorithm-level generalization.

\subsection{DNA tasks and oracles}
We use the following three DNA oracles which predict chromatin accessibility in K562. They differ in assay (ATAC vs.\ DNase), input length, and architecture, so a sequence that exploits one model's inductive bias need not score well on the others.

\paragraph{BPNet ATAC.}
Chromatin accessibility measures how available DNA is to regulatory proteins and is a widely used readout of gene-regulatory activity. We use the NucleoBench BPNet-lite ATAC model~\parencite{bpnet-lite,Shor2025NucleoBench}, a dilated residual CNN~\parencite{bpnet} trained on ENCODE K562 ATAC-seq~\parencite{encode}. Designers optimize a 3{,}000~bp window.

\paragraph{ChromBPNet.}
The same accessibility phenotype, with explicit correction for Tn5 cut bias. We use a ChromBPNet model trained on K562 ATAC-seq (ENCSR483RKN)~\parencite{Pampari2024ChromBPNet}: a fully convolutional network of dilated residual layers with a frozen background bias model whose contribution is subtracted from the accessibility head. Inputs are 2{,}114~bp.

\paragraph{Enformer.}
A long-range sequence-to-activity transformer that predicts thousands of human and mouse epigenetic and transcriptional tracks from 196{,}608~bp of DNA~\parencite{enformer}. For this panel we score K562 DNase tracks, aggregated over the 128~bp bins that overlap the central BPNet window, so the readout is accessibility in the same cell type rather than NucleoBench's muscle-versus-liver expression task.

\begin{figure}[ht]
    \centering
    \includegraphics[width=\columnwidth]{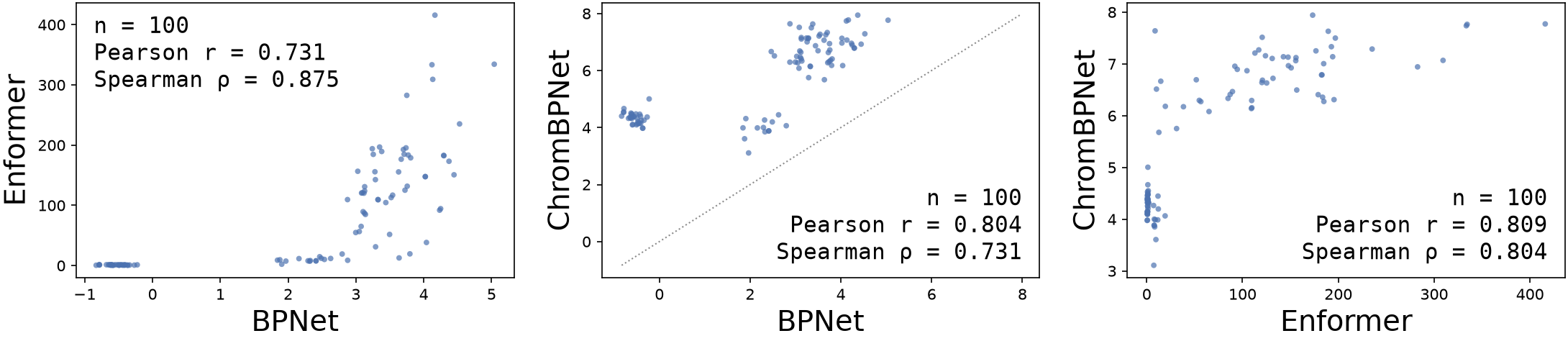}
    \caption{Pairwise oracle scores on the same 100 NucleoBench start sequences across the DNA panel (BPNet ATAC, Enformer K562 DNase, ChromBPNet K562).}
    \label{fig:panel1}
\end{figure}

\textbf{Correlated but still different.} \Cref{fig:panel1} scores the same $N{=}100$ NucleoBench start sequences under every pairwise combination of DNA oracles. DNA oracles correlate substantially but not perfectly (Pearson $r \approx 0.73$--$0.81$). Imperfect correlation is the point: a designer that overfits one oracle need not improve the others.

\section{Results}\label{sec:results}
We compare four baselines (AdaBeam, GrAdaBeam, FastSeqProp, and Ledidi) with sixteen AutoNovo-discovered designers under a 30-minute sequence-design budget. All values are mean $\pm$ standard deviation NucleoBench oracle scores over $N{=}20$ start sequences unless otherwise indicated, and lower is better. One AutoNovo candidate ran out of memory when Enformer was the design oracle, so its Enformer direct result and its two Enformer-origin transfer results are unavailable. The AutoNovo designers in the final evaluation were derived from FastSeqProp, Ledidi, and PEX. ERA also attempted to reimplement and improve GrAdaBeam, but failed to produce a version competitive with the original GrAdaBeam implementation; we therefore report GrAdaBeam only as a baseline.

\paragraph{Direct performance.}
\Cref{tab:direct_tasks} reports performance when the design and evaluation oracles are the same. AutoNovo designers achieve the best observed mean on all three tasks. The best AutoNovo designers score $-61.17$ on BPNet ATAC, $-74.88$ on ChromBPNet, and $-3360.36$ on Enformer, compared with the strongest baseline means of $-40.88$, $-60.08$, and $-124.53$, respectively. The best designer differs across tasks: FSP-gradient memory STE performs best on BPNet ATAC, FSP-Best-Anchored STE on ChromBPNet, and FSP-sparse restart STE on Enformer. Improvements are not confined to a single selected candidate: 13 of 16 AutoNovo designers outperform the strongest baseline mean on BPNet ATAC, 7 of 16 do so on ChromBPNet, and 11 of the 15 completed candidates do so on Enformer.

\begin{table}[htbp]
\centering
\caption{Direct performance of sequence design after 30 minutes on CPU. Each cell reports the mean on the first line and standard deviation on the second across $N=20$ start sequences; lower is better. AutoNovo designers are grouped by the optimization oracle $f_{\mathrm{opt}}$ and orthogonal oracle $f_{\mathrm{orth}}$ used during ERA search. The FSP, Ledidi, and PEX prefixes identify the seed algorithm. Bold indicates the best observed mean in each column.}
\label{tab:direct_tasks}
\renewcommand{\arraystretch}{1.1}
\setlength{\tabcolsep}{4pt}
\small
\begin{tabular}{lccc}
\toprule
\textbf{Designer}
    & \textbf{\texttt{bpnet}}
    & \textbf{\texttt{chrombpnet}}
    & \textbf{\texttt{enformer}} \\
\midrule

\multicolumn{4}{l}{\textbf{Baselines}} \\
\texttt{adabeam}
    & \meanerr{-21.90}{3.53}
    & \meanerr{-37.71}{6.91}
    & \meanerr{-61.28}{73.78} \\
\texttt{fastseqprop}
    & \meanerr{-23.27}{5.64}
    & \meanerr{-46.26}{6.24}
    & \textit{OOM} \\
\texttt{gradabeam}
    & \meanerr{-27.85}{3.26}
    & \meanerr{-38.42}{4.86}
    & \meanerr{-62.67}{75.32} \\
\texttt{ledidi}
    & \meanerr{-40.88}{2.58}
    & \meanerr{-60.08}{4.53}
    & \meanerr{-124.53}{148.27} \\

\midrule
\multicolumn{4}{l}{\textbf{AutoNovo-discovered designers}} \\

\addlinespace
\multicolumn{4}{l}{
    \scriptsize
    \textit{ERA search:}
    $f_{\mathrm{opt}}$ = BPNet ATAC;
    $f_{\mathrm{orth}}$ = ChromBPNet
} \\
\texttt{FSP-noisy reset softmax}
    & \meanerr{-37.73}{4.37}
    & \meanerr{-41.27}{5.85}
    & \meanerr{-443.11}{454.37} \\
\texttt{FSP-best steered batch search}
    & \meanerr{-38.09}{2.34}
    & \meanerr{-37.83}{5.47}
    & \meanerr{-821.45}{237.18} \\
\texttt{Ledidi-windowed STE}
    & \meanerr{-49.61}{2.61}
    & \meanerr{-42.31}{4.08}
    & \meanerr{-229.95}{234.53} \\

\addlinespace
\multicolumn{4}{l}{
    \scriptsize
    \textit{ERA search:}
    $f_{\mathrm{opt}}$ = ChromBPNet;
    $f_{\mathrm{orth}}$ = BPNet ATAC
} \\
\texttt{FSP-gradient centering}
    & \meanerr{-45.47}{2.38}
    & \meanerr{-66.20}{4.88}
    & \meanerr{-1166.94}{409.19} \\
\texttt{FSP-sparse restart STE}
    & \meanerr{-53.05}{1.21}
    & \meanerr{-54.40}{5.28}
    & \bestmeanerr{-3360.36}{296.97} \\
\texttt{Ledidi-length gated hybrid}
    & \meanerr{-45.31}{2.18}
    & \meanerr{-68.48}{3.17}
    & \meanerr{-85.23}{102.14} \\

\addlinespace
\multicolumn{4}{l}{
    \scriptsize
    \textit{ERA search:}
    $f_{\mathrm{opt}}$ = BPNet ATAC;
    $f_{\mathrm{orth}}$ = Enformer
} \\
\texttt{FSP-diverse batch STE}
    & \meanerr{-52.19}{1.29}
    & \meanerr{-58.03}{3.37}
    & \textit{OOM} \\
\texttt{FSP-gradient memory STE}
    & \bestmeanerr{-61.17}{1.52}
    & \meanerr{-70.80}{3.69}
    & \meanerr{-1209.13}{744.51} \\
\texttt{Ledidi-perturbed STE}
    & \meanerr{-46.98}{2.63}
    & \meanerr{-56.18}{3.42}
    & \meanerr{-1363.04}{274.47} \\

\addlinespace
\multicolumn{4}{l}{
    \scriptsize
    \textit{ERA search:}
    $f_{\mathrm{opt}}$ = ChromBPNet;
    $f_{\mathrm{orth}}$ = Enformer
} \\
\texttt{FSP-gradient resync}
    & \meanerr{-41.73}{2.87}
    & \meanerr{-57.06}{6.32}
    & \meanerr{-59.40}{73.77} \\
\texttt{FSP-best-anchored STE}
    & \meanerr{-49.85}{2.41}
    & \bestmeanerr{-74.88}{2.29}
    & \meanerr{-2614.21}{400.08} \\
\texttt{Ledidi-greedy gradient hybrid}
    & \meanerr{-43.09}{4.02}
    & \meanerr{-61.79}{6.73}
    & \meanerr{-111.78}{126.24} \\

% added for PEX
\addlinespace
\multicolumn{4}{l}{
    \scriptsize
    \textit{ERA search:}
    $f_{\mathrm{opt}}$ = BPNet ATAC + ChromBPNet; no hidden oracle
} \\
\texttt{FSP-Ledidi-PEX two-phase saliency}
    & \meanerr{-35.16}{2.26}
    & \meanerr{-24.47}{3.12}
    & \meanerr{-83.01}{97.41} \\
\texttt{FSP-Ledidi-PEX annealed STE}
    & \meanerr{-51.89}{1.76}
    & \meanerr{-67.06}{3.43}
    & \meanerr{-871.53}{336.48} \\
\texttt{FSP-Ledidi-PEX stall-tempered STE}
    & \meanerr{-46.90}{2.30}
    & \meanerr{-41.22}{3.87}
    & \meanerr{-2955.35}{364.66} \\
\texttt{FSP-Ledidi-PEX latency-gated STE}
    & \meanerr{-53.81}{1.37}
    & \meanerr{-61.47}{4.00}
    & \meanerr{-131.76}{166.95} \\
% added for PEX // end

\bottomrule
\end{tabular}
\end{table}

\paragraph{Cross-oracle transfer.}
\Cref{tab:transfer_task_results} reports the score assigned by an evaluation oracle different from the oracle used for sequence design. An AutoNovo designer achieves the best observed mean in all six transfer directions, although no single candidate wins every direction. Improvements are especially consistent for BPNet ATAC $\rightarrow$ Enformer, where 15 of 16 AutoNovo designers outperform the strongest baseline mean, and for the Enformer-origin transfers, where 13 of 15 completed candidates do so on Enformer $\rightarrow$ BPNet ATAC and 12 of 15 on Enformer $\rightarrow$ ChromBPNet. Performance is less uniform when ChromBPNet is the design oracle: 4 of 16 candidates outperform the strongest baseline on ChromBPNet $\rightarrow$ BPNet ATAC, and 4 of 16 do so on ChromBPNet $\rightarrow$ Enformer. Thus, the results support a family of effective AutoNovo-discovered designers, while also showing meaningful specialization among the generated algorithms.

\begin{table}[htbp]
\centering
\caption{Cross-oracle transfer performance after 30 minutes of sequence design. $A \rightarrow B$ denotes sequences designed against oracle $A$ and evaluated by oracle $B$ at test time. Each cell reports the mean on the first line and standard deviation on the second across $N=20$ start sequences; lower is better. AutoNovo designers are grouped by the optimization oracle $f_{\mathrm{opt}}$ and orthogonal oracle $f_{\mathrm{orth}}$ used during ERA search. The FSP, Ledidi, and PEX prefixes identify the seed algorithm. Bold indicates the best observed mean in each column, not statistical significance.}
\label{tab:transfer_task_results}
\renewcommand{\arraystretch}{1.1}
\setlength{\tabcolsep}{2pt}
\scriptsize
\resizebox{\linewidth}{!}{%
\begin{tabular}{lcccccc}
\toprule
\textbf{Designer}
    & \makecell{\textbf{\texttt{bpnet}} \\
      $\boldsymbol{\rightarrow}$ \textbf{\texttt{chrom}}}
    & \makecell{\textbf{\texttt{bpnet}} \\
      $\boldsymbol{\rightarrow}$ \textbf{\texttt{enformer}}}
    & \makecell{\textbf{\texttt{chrom}} \\
      $\boldsymbol{\rightarrow}$ \textbf{\texttt{bpnet}}}
    & \makecell{\textbf{\texttt{chrom}} \\
      $\boldsymbol{\rightarrow}$ \textbf{\texttt{enformer}}}
    & \makecell{\textbf{\texttt{enformer}} \\
      $\boldsymbol{\rightarrow}$ \textbf{\texttt{bpnet}}}
    & \makecell{\textbf{\texttt{enformer}} \\
      $\boldsymbol{\rightarrow}$ \textbf{\texttt{chrom}}} \\
\midrule

\multicolumn{7}{l}{\textbf{Baselines}} \\
\texttt{adabeam}
    & \meanerr{-10.05}{1.25}
    & \meanerr{-803.77}{142.81}
    & \meanerr{-4.45}{0.39}
    & \meanerr{-502.41}{110.40}
    & \meanerr{-1.52}{1.95}
    & \meanerr{-5.37}{1.24} \\
\texttt{fastseqprop}
    & \meanerr{-10.18}{0.90}
    & \meanerr{-711.98}{321.24}
    & \meanerr{-6.48}{0.61}
    & \meanerr{-701.38}{183.28}
    & \textit{OOM}
    & \textit{OOM} \\
\texttt{gradabeam}
    & \meanerr{-10.47}{1.18}
    & \meanerr{-931.81}{147.39}
    & \meanerr{-4.57}{0.43}
    & \meanerr{-483.59}{174.38}
    & \meanerr{-1.53}{1.93}
    & \meanerr{-5.38}{1.23} \\
\texttt{ledidi}
    & \meanerr{-9.99}{0.83}
    & \meanerr{-1054.55}{184.66}
    & \meanerr{-7.00}{0.81}
    & \meanerr{-937.19}{233.70}
    & \meanerr{-1.73}{1.83}
    & \meanerr{-5.52}{1.29} \\

\midrule
\multicolumn{7}{l}{\textbf{AutoNovo-discovered designers}} \\

\addlinespace
\multicolumn{7}{l}{
    \scriptsize
    \textit{ERA search:}
    $f_{\mathrm{opt}}$ = BPNet ATAC;
    $f_{\mathrm{orth}}$ = ChromBPNet
} \\
\texttt{FSP-noisy reset softmax}
    & \meanerr{-10.84}{1.35}
    & \meanerr{-1105.23}{166.54}
    & \meanerr{-4.63}{0.56}
    & \meanerr{-558.66}{124.64}
    & \meanerr{-3.60}{1.04}
    & \meanerr{-7.27}{0.58} \\
\texttt{FSP-best steered batch search}
    & \meanerr{-11.13}{1.20}
    & \meanerr{-1076.14}{169.61}
    & \meanerr{-4.26}{0.53}
    & \meanerr{-355.05}{171.95}
    & \meanerr{-4.58}{0.36}
    & \meanerr{-7.74}{0.37} \\
\texttt{Ledidi-windowed STE}
    & \meanerr{-9.81}{0.86}
    & \meanerr{-1114.38}{138.64}
    & \meanerr{-5.84}{0.55}
    & \meanerr{-666.57}{191.86}
    & \meanerr{-2.64}{1.41}
    & \meanerr{-6.09}{1.34} \\

\addlinespace
\multicolumn{7}{l}{
    \scriptsize
    \textit{ERA search:}
    $f_{\mathrm{opt}}$ = ChromBPNet;
    $f_{\mathrm{orth}}$ = BPNet ATAC
} \\
\texttt{FSP-gradient centering}
    & \meanerr{-10.86}{1.16}
    & \meanerr{-1115.76}{103.25}
    & \meanerr{-5.65}{0.75}
    & \meanerr{-799.27}{204.50}
    & \meanerr{-4.59}{0.45}
    & \meanerr{-7.81}{0.60} \\
\texttt{FSP-sparse restart STE}
    & \meanerr{-9.96}{0.68}
    & \meanerr{-1126.64}{133.46}
    & \meanerr{-6.54}{0.76}
    & \meanerr{-713.78}{138.38}
    & \bestmeanerr{-5.35}{0.38}
    & \bestmeanerr{-8.27}{0.41} \\
\texttt{Ledidi-length gated hybrid}
    & \meanerr{-9.97}{1.05}
    & \meanerr{-1068.25}{143.37}
    & \meanerr{-7.11}{0.57}
    & \meanerr{-931.05}{175.07}
    & \meanerr{-1.81}{1.77}
    & \meanerr{-5.62}{1.29} \\

\addlinespace
\multicolumn{7}{l}{
    \scriptsize
    \textit{ERA search:}
    $f_{\mathrm{opt}}$ = BPNet ATAC;
    $f_{\mathrm{orth}}$ = Enformer
} \\
\texttt{FSP-diverse batch STE}
    & \meanerr{-10.03}{0.88}
    & \meanerr{-1239.54}{177.55}
    & \meanerr{-6.72}{0.63}
    & \meanerr{-1004.37}{152.25}
    & \textit{OOM}
    & \textit{OOM} \\
\texttt{FSP-gradient memory STE}
    & \meanerr{-9.06}{0.80}
    & \meanerr{-1098.01}{126.21}
    & \bestmeanerr{-7.63}{0.61}
    & \meanerr{-822.73}{241.38}
    & \meanerr{-4.43}{0.74}
    & \meanerr{-7.58}{0.81} \\
\texttt{Ledidi-perturbed STE}
    & \meanerr{-9.70}{0.81}
    & \meanerr{-1067.93}{166.73}
    & \meanerr{-6.51}{0.70}
    & \meanerr{-892.29}{141.73}
    & \meanerr{-4.26}{0.44}
    & \meanerr{-7.57}{0.63} \\

\addlinespace
\multicolumn{7}{l}{
    \scriptsize
    \textit{ERA search:}
    $f_{\mathrm{opt}}$ = ChromBPNet;
    $f_{\mathrm{orth}}$ = Enformer
} \\
\texttt{FSP-gradient resync}
    & \meanerr{-10.46}{1.09}
    & \meanerr{-1103.15}{166.20}
    & \meanerr{-5.21}{0.60}
    & \meanerr{-799.82}{205.09}
    & \meanerr{-1.51}{2.00}
    & \meanerr{-5.37}{1.27} \\
\texttt{FSP-best-anchored STE}
    & \meanerr{-9.45}{1.24}
    & \meanerr{-1092.26}{171.54}
    & \meanerr{-7.39}{0.60}
    & \bestmeanerr{-1071.39}{165.09}
    & \meanerr{-4.67}{0.39}
    & \meanerr{-7.96}{0.59} \\
\texttt{Ledidi-greedy gradient hybrid}
    & \meanerr{-9.87}{0.74}
    & \meanerr{-973.75}{162.39}
    & \meanerr{-6.75}{0.64}
    & \meanerr{-868.64}{164.91}
    & \meanerr{-1.84}{1.98}
    & \meanerr{-5.69}{1.39} \\

% added for PEX
\addlinespace
\multicolumn{7}{l}{
    \scriptsize
    \textit{ERA search:}
    $f_{\mathrm{opt}}$ = BPNet ATAC + ChromBPNet; no hidden oracle
} \\
\texttt{FSP-Ledidi-PEX two-phase saliency}
    & \bestmeanerr{-11.58}{1.14}
    & \meanerr{-1096.51}{129.68}
    & \meanerr{-4.15}{0.29}
    & \meanerr{-152.75}{134.43}
    & \meanerr{-1.68}{1.83}
    & \meanerr{-5.50}{1.23} \\
\texttt{FSP-Ledidi-PEX annealed STE}
    & \meanerr{-11.04}{1.22}
    & \bestmeanerr{-1303.44}{141.56}
    & \meanerr{-6.38}{0.57}
    & \meanerr{-970.34}{226.94}
    & \meanerr{-4.25}{0.51}
    & \meanerr{-7.63}{0.82} \\
\texttt{FSP-Ledidi-PEX stall-tempered STE}
    & \meanerr{-9.96}{0.72}
    & \meanerr{-1056.82}{145.87}
    & \meanerr{-5.98}{0.74}
    & \meanerr{-685.20}{138.47}
    & \meanerr{-4.96}{0.26}
    & \meanerr{-8.22}{0.67} \\
\texttt{FSP-Ledidi-PEX latency-gated STE}
    & \meanerr{-9.56}{0.56}
    & \meanerr{-1159.48}{144.17}
    & \meanerr{-7.02}{0.48}
    & \meanerr{-1026.79}{148.17}
    & \meanerr{-1.81}{1.83}
    & \meanerr{-5.49}{1.31} \\
% added for PEX // end

\bottomrule
\end{tabular}%
}
\end{table}

\textbf{Deviation from NucleoBench results for baseline designers.} The rank-order performance of baseline designers is different in this study than in NucleoBench. The primary cause is the time budget: NucleoBench ran for 8 hours, while this study ran for 30 minutes (both on CPU). Ledidi designs quickly as compared to Fast SeqProp, AdaBeam, and GrAdaBeam. The absolute numbers for all designers are worse both because of the shorter design time, and the hyperparameters were selected from the NucleoBench 8-hour runs.

\section{What Algorithms Did AutoNovo Discover}

Overall, AutoNovo discovered a family of optimizers that explicitly took advantage of time/compute budget and successfully blended strategies from other designers. Their most consistent ingredients are hard-sequence gradient optimization, explicit preservation of the best sequence, and relatively simple mechanisms for restarting search after temporarily slowing progress.

\paragraph{The best programs use a hybrid of continuous and discrete search.}
Most successful designers optimize continuous sequence logits with gradients, convert those logits into discrete sequences, retain the best discrete sequence found, and perturb the continuous state when progress stalls. Every best observed mean in \cref{tab:direct_tasks,tab:transfer_task_results} comes from a FastSeqProp-seeded program. Eight of the nine winning table entries come from programs using a straight-through estimator (STE), which evaluates a hard one-hot sequence while passing gradients through a soft relaxation.

The strongest programs implement this common pattern in different ways. FSP-Best-Anchored STE, which performs best on direct ChromBPNet and ChromBPNet $\rightarrow$ Enformer, periodically pulls its logits toward the best sequence found so far. FSP-gradient memory STE, which performs best on direct BPNet ATAC and ChromBPNet $\rightarrow$ BPNet ATAC, accumulates gradient magnitudes and uses them to make guided jumps after stalling. FSP-sparse restart STE, which performs best on direct Enformer and both Enformer-origin transfer tasks, restarts around a sparsely mutated version of its best sequence. In each case, gradients guide the search through discrete sequence space.

\paragraph{Designers often performed best on the oracle pairs used during their search, and some generalized to other tasks.}
One of the four transfer objectives used during ERA search is won by a program searched on the corresponding oracle pair: FSP-Best-Anchored STE for ChromBPNet $\rightarrow$ Enformer. Both BPNet-origin directions are instead won by programs searched against BPNet ATAC and ChromBPNet together.
The generated program cannot query the orthogonal oracle directly, but ERA uses its score when selecting programs. This lets the outer loop favor algorithms whose sequences transfer even though the inner designer sees only its optimization oracle.

\paragraph{Two-oracle searches produced both BPNet-origin transfer winners.}
Four designers were searched against BPNet ATAC and ChromBPNet together. ``FSP-Ledidi-PEX annealed STE'' has the best observed mean in BPNet ATAC $\rightarrow$ Enformer, even though Enformer was queried by neither the designer nor ERA at any point in its search. This improvement transfers to an evaluation oracle absent from the entire program-search loop. The decoupling runs the other way too. ``FSP-Ledidi-PEX two-phase saliency'' has the weakest direct BPNet ATAC score of any AutoNovo designer, yet it wins BPNet ATAC $\rightarrow$ ChromBPNet. Scoring well against the design oracle and producing sequences that survive an independent oracle are not the same objective, which supports this paper's hypothesis that careful selection of $f_{\mathrm{opt}}$ and $f_{\mathrm{orth}}$ can improve generalization.

No program is best everywhere, but some programs transfer beyond the configuration in which ERA discovered them. \textbf{The best example of a generalizing designer is FSP-sparse restart STE}: ERA searched for it using ChromBPNet as the optimization oracle and BPNet ATAC as the orthogonal evaluator, yet it is the best program when Enformer is used for design. It also wins both Enformer-origin transfer tasks. This is evidence that ERA discovered an optimization strategy that transfers to an oracle and sequence length not used for its inner-loop optimization during search.

\paragraph{On Enformer (large model), memory-efficiency outperforms large candidate batches.}
FSP-diverse batch STE generates a batch of roughly thirty candidates during discrete search. This designer performs second-best on BPNet ATAC $\rightarrow$ Enformer. The same designer runs out of memory when Enformer becomes the design oracle, due to the longer sequences.

FSP-sparse restart STE takes the opposite approach. It follows a single STE trajectory, evaluates one discrete sequence at a time, checks candidates more frequently near the deadline, and uses sparse mutations after stalling. It is the strongest Enformer designer. The comparison suggests that under a fixed wall-clock and memory budget, candidate batching and evaluation frequency strongly determine scaling properties.

\paragraph{Different seed algorithms converge on similar mechanisms.}
The Ledidi-seeded programs also moved toward the same continuous--discrete design. The Ledidi description given to ERA emphasizes an explicit $L_1$ edit-distance penalty, but none of the four generated programs had one. Instead, they use the same general ingredients as the FastSeqProp-seeded programs: logits, Adam, STEs, best-sequence tracking, discrete mutations, and recovery after stagnation.

The seeds therefore acted as loose priors rather than constraints on the final algorithms. All seed families converged toward a similar set of mechanisms. This could mean that the benchmark strongly favors this hybrid design, that the language model has a strong prior for it, or both. The prompt also explicitly suggested adapting gradient use to task and runtime, so the broad idea of adaptive gradient use was supplied to ERA.

\paragraph{Strong scores did not imply clean or reliable code.}
Several generated programs use sequence length as a proxy for oracle identity and change their use of gradients accordingly. One Ledidi-seeded program disables gradients on Enformer-length inputs and performs poorly there. Other programs contain unused controls, continue using stale gradients after switching away from gradient optimization, or include comments that do not match the executed condition. None sets random seeds. These defects do not prevent the program family from performing strongly, but they show why agent-generated scientific code still requires inspection, reproducibility checks, and targeted ablations.

\section{Agent and Harness Design Observations}\label{sec:factors}

The following are strategies we found in our AutoNovo experiments to improve the ability of ERA to generate correct designers:

\paragraph{Ban \texttt{try}/\texttt{except}.}
LLM-written designers often wrapped buggy code in \texttt{try}/\texttt{except} blocks, which propagated broken code. To get it to write code that worked and was written cleanly, our prompt forbids exception handlers, and the harness AST-checks the \texttt{Designer} class: any \texttt{try} with a handler receives the worst-case score. Crashes then surfaced and were scored as failures (correctly).

\paragraph{Seed with NucleoBench algorithms.}
Initial prompts asked ERA to ``improve a designer with gradient information''. While these worked reasonably fine, the performances were subpar relative to the top hand-crafted designer. Subsequently we included detailed descriptions of existing designers GrAdaBeam~\parencite{Shor2025NucleoBench}, FastSeqProp~\parencite{fastseqprop}, Ledidi~\parencite{ledidi}, and Proximal Exploration~\parencite{pex}, each in separate runs which resulted in substantially improved performance. We also added the insight from NucleoBench that gradients help on some tasks (GATA2, ATAC) and hurt on others (JUNB, OTX1), possibly because they are uninformative or too slow. This information, along with a note to generalize helped the system match or exceed performance of the chosen original algorithm.

\paragraph{Pass the wall-clock budget and require that it be used.}
NucleoBench evaluates designers on time, not step count. Results showed that algorithms which \emph{see} the remaining budget outperform those that do not. The interface was adapted to accept \texttt{run(allowed\_runtime\_secs)}. The prompt states this explicitly and asks the designer to meaningfully use it.

\paragraph{Tell the designer that gradients are optional and costly.}
Because evaluation is itself a time-bounded optimization, a backward pass that does not significantly improve performance is harmful. Prompts state that gradient usefulness varies by task and even by phase of a single run, and that a design goal is to decide \emph{when} to use them.

\paragraph{Make failures expensive, and show how to take gradients.}
Timeouts, crashes, and the \texttt{try}/\texttt{except} ban all map to a large penalty, to discourage them. Errors must be as informative as possible. Separately, the notebook includes working PyTorch gradient snippets for each oracle. Those cells are not the rewrite target; they exist so ERA does not spend generations on framework code.

\section{Conclusion}\label{sec:conclusion}

This study explores whether agentic program search can improve nucleic acid design algorithms. Across three chromatin-accessibility oracles, an AutoNovo-discovered designer achieved the best observed mean on each of the three direct optimization tasks and all six cross-oracle transfer directions. Multiple AutoNovo designers also surpassed the strongest baseline on several tasks, although different generated programs performed best on different oracles. We also identified practical factors that improved ERA’s ability to generate useful designers, including providing strong algorithmic priors, exposing the available wall-clock budget, and allowing the designer to decide when gradients are worth their computational cost.

These results show that agentic program search can produce scientific optimization algorithms whose improvements can extend beyond the oracle pair and optimization configuration used during program discovery. However, the evidence remains limited to in-silico evaluation of one biological phenotype using three related predictive models, and cross-oracle transfer does not establish wet-lab validity. Future work should test AutoNovo on additional biological design problems, determine which algorithmic changes account for successful transfer, and evaluate whether cross-oracle improvements predict performance in biological experiments.

\printbibliography

\newpage

\appendix
\section{Appendix}
\begin{promptbox}[System Prompt]
# Improving a nucleic acid designer potentialy with gradient info.

The goal of the task is to improve an existing designer that takes in a nucleic acid string and iteratively modifies it to maximize its fitness (minimize oracle prediction loss/score) according to a pretrained oracle.  We want to create designers or improve existing designers so that they **generalize to many different tasks**.

For any starting sequence, write the `__init__`, `run`, and `get_samples` methods of the `Designer` class such that the starting sequence (`start_sequence`) is iteratively modified so as to minimize its `model_fn` score. If `seq_n` is the current modification of the sequence, the goal is to minimize `self.model_fn([seq_n])[0]`.

You will be evaluated on an independent evaluation oracle that has learned about the same underlying biological phenomenon.

Constraints & Details:
- Vocabulary: `['A', 'C', 'G', 'T']`
- `seq_n` must be exactly the same length (`self.seq_len`) as the starting sequence.
- The objective is to **minimize** the candidate metric.

**NOTE** 
1. Make sure to print out logs along the way to ensure you have loaded the models correctly and are able to compute gradients.
2. Do not use try-except clauses in the code. This will help ensure the behavior is as expected and not going into the except clauses.


### Additional problem notes

Some tasks can use gradients to great effect, but some tasks they appear detrimental. This could be the case because gradients are not very informative in some problems, or it could be because they simply aren't worth the time it takes to compute them (remember, the evaluation is an optimization problem, and thus time-per-compute is an important feature). The **task gradients have different properties**. For example, gradient methods do well on GATA2, CTCF, MAC, ATAC. They do especially poorly on Rinalmo, JUNB, and OTX1. Furthermore, gradients might be useful at some points in the design phase, and might become less useful in other points. One of your algorithmic design goals will be to adaptively determine if and how to use gradients in a time-efficient manner (remembering that they are relatively expensive to compute and may be useful at times but not always). so we show you how to take gradients from different models but you can explore and decide based on trying what leads to improvement.

\end{promptbox}

\end{document}